\documentclass[12pt]{dalthesis}
\usepackage{hyperref}
\usepackage{color}
\usepackage{graphicx}
\usepackage{booktabs}
\usepackage{subfigure}
\usepackage{listings}
\usepackage{multirow}
\usepackage{mathtools}
\usepackage{amsmath}
\usepackage{algorithm}
\usepackage{algorithmic}
\begin{document}

\title{Transportation Modes Classification Using Feature Engineering}
\author{Mohammad Etemad}

\phd  


\degree{PhD of Computer Science}
\degreeinitial{PhD}
\faculty{Computer Science}
\dept{Faculty of Computer Science}

\defencemonth{Sep}\defenceyear{2018}



\frontmatter

\begin{abstract}
Predicting transportation modes from GPS (Global Positioning System) records is a hot topic in the trajectory mining domain. Each GPS record is called a trajectory point and a trajectory is a sequence of these points.
Trajectory mining has applications including but not limited to 
transportation mode detection, tourism, traffic congestion, smart cities management, animal behaviour analysis, environmental preservation, and
traffic dynamics are some of the trajectory mining applications.
Transportation modes prediction as one of the tasks in human mobility and vehicle mobility applications plays an important role in resource allocation, traffic management systems, tourism planning and accident detection.
In this work, the proposed framework in \cite{etemad2018predicting} is extended to consider other aspects in the task of transportation modes prediction. 

Wrapper search and information retrieval methods were investigated to find the best subset of trajectory features.
Finding the best classifier and the best feature subset, the framework is compared against two related papers that applied deep learning methods. The results show that our framework achieved better performance. Moreover, the ground truth noise removal improved accuracy of transportation modes prediction task; however, the assumption of having access to test set labels in pre-processing task is invalid. Furthermore, the cross validation approaches were investigated and the performance results show that the random cross validation method provides optimistic results.

\end{abstract}



\mainmatter

\chapter{Introduction}

Trajectory mining is a very hot topic since positioning devices are now used to track people, vehicles, vessels, natural phenomena and animals. 
It has applications including but not limited to transportation mode detection \cite{zheng2010understanding,endo2016deep,dabiri2018inferring,xiao2017identifying}, fishing detection\cite{de2016improving,hu2016identifying}, tourism \cite{Parent:2013:STM:2501654.2501656,feng2017poi2vec}, 
animal behaviour analysis \cite{jonsen2007identifying,fossette2010spatio}, 
climate science \cite{karpatne2013earth,doi:10.1175/1520-0477(2001)082<0247:TNNYRM>2.3.CO;2},
neuroscience \cite{atluri2016brain},
environmental science \cite{thompson2014systems,dividinosemantic},
precision agriculture \cite{mahlein2016plant},
epidemiology and health care \cite{matsubara2014funnel},
social media \cite{carney2016all},
traffic dynamics \cite{castro2013taxi},
heliophysics \cite{Hurlburt2012}, and crime data \cite{bappee2018predicting,tompson2015uk}.
Human mobility and vehicle mobility, as a small subset of the wide range of trajectory mining applications, can be used in resource allocation, traffic management systems, tourism planning and accident detection \cite{amin2012accident,castro2013taxi,Parent:2013:STM:2501654.2501656,feng2017poi2vec,jiang2017trajectorynet}.

There are a number of open topics in this field that need to be investigated further such as high performance trajectory classification methods \cite{endo2016deep,dabiri2018inferring,zheng2010understanding,xiao2017identifying,liu2017end}, accurate trajectory segmentation methods \cite{zheng2008understanding,yan2011setrastream,yan2011setrastream,soares2015grasp,hwang2018segmenting}, trajectory compression and reduction \cite{sun2016overview,hershberger1992speeding}, privacy in trajectory mining \cite{chen2013privacy,ho2011differential,giannotti2008mobility}, trajectory similarity and clustering \cite{kang2009similarity,fu2005similarity,nanni2006time,kisilevich2009spatio}, dealing with trajectory uncertainty \cite{jun2006smoothing,hwang2018segmenting}, and semantic trajectories \cite{parent2013semantic,klotz2018generating,moreno2015}, and active learning strategies for trajectory data \cite{alearning2017}. 
These topics are highly correlated and solving one of them requires to some extent exploring the other questions. 
For example, performing trajectory classification needs to deal with noise and segmentation directly and the other aforementioned topics indirectly. 

It is important to point out that the heart of the trajectory prediction task is the data itself and the accurate capture of raw trajectory records. 
Some enriched sources of raw trajectories are available in the public domain but the majority of them are proprietary. 
For example, the GeoLife GPS Trajectory\footnote{https://www.microsoft.com/en-us/download/details.aspx?id=52367} Dataset is a publicly available raw trajectory dataset collected by Microsoft Research from April 2007 to August 2012\cite{zheng2011geolife}. 
This dataset is applied for evaluation in many research studies such as \cite{dabiri2018inferring,xiao2017identifying,zheng2010understanding,endo2016deep,liu2017end}. Another publicly available dataset is T-Drive Taxi Trajectories\footnote{https://www.microsoft.com/en-us/research/publication/t-drive-trajectory-data-sample/} that was also collected by Microsoft Research from March 2009 to August 2009. 
The Hurricanes Dataset\footnote{https://www.nhc.noaa.gov/data/} is another trajectory dataset publicly available provided by the National Hurricane Service (NHS) from 1851 to 2012. 
The Movebank\footnote{https://www.movebank.org/} Animal trajectory is another publicly available source of raw trajectories .
As more examples, The Automatic Identification System (AIS) and  Satellite-AIS (S-AIS) Dataset are other very large scale trajectory datasets developed to monitor vessels worldwide. 
These datasets are public resources, but the antennas/satellites to collect the data are not.
Some processed samples of AIS data are available from 2009 to 2014 and published under the name of Vessel Traffic Data\footnote{https://marinecadastre.gov/ais/}  by National Oceanic and Atmospheric Administration(NOAA).

The Geolife dataset and the trajectory classification task were selected as the focus of this research to start exploring this field. 
Therefore, the related works were selected from papers that investigate the transportation mode classification using the GeoLife dataset. 
Using two feature selection approaches, it was investigated  the best subset of features for transportation modes prediction. 
Furthermore, using the best classifier and the best subset of features, the results were compared with the works of \cite{dabiri2018inferring,endo2016deep} and the results showed that our approach achieved a better result. 
Finally, this work investigated the differences between two methods of cross validation and the results show that the random cross validation method suggests optimistic results in comparison to user oriented cross validation.

The rest of this work is structured as follows. 
The related works are reviewed in chapter \ref{sec:PreviousWork}. 
The basic concepts and definitions are provided in chapter \ref{sec:Preliminaries}. 
The Geolife dataset is described in section \ref{sec:GeoLifeDataset}. 
Section \ref{sec:UncertaintyofData} talks about handling noise and harnessing the uncertainty of data. 
The applied framework is detailed in section \ref{sec:model}. 
We provide our experimental results in chapter 3. Finally, we conclude the report in chapter 4.

\chapter{Related works}
\label{sec:PreviousWork}
In this section, some recent research on the trajectory classification task were analyzed using the Geolife dataset. 
A trajectory is a sequence of GPS points captured through time. 
We define a formal definition of a trajectory in Chapter 3.

Four important aspects of related research include: first, the computational resources needed for training, like use of CPU or GPU; second, the features were used for training their model such as representation learning or hand crafted features; third, their evaluation methods such as different ways of doing cross validation; and forth, noise removal approaches like smoothing or ground truth. 
Firstly, an overall review of each paper is provided. 
Then, each aspect is discussed.

Zheng et al. (2008) conducted a study to recognize human behavior and understand users' mobility. 
The user behaviour analyzed in this work included the transportation means walking, driving, bus, and bike - four classes \cite{zheng2008understanding}. 
This supervised learning approach is the marriage of generating sophisticated features and a graph-based post processing algorithm for improving the prediction performance of transportation modes. 

The proposed model by Zheng et al.\cite{zheng2010understanding} is evaluated using the Geolife dataset as follows. 
First, a segmentation method, explained in \cite{Zheng2008}, builds on the concepts that people must stop when they switch from one transportation mode to another; walking behaviour happens between each two other transition modes. 
After, each user's trajectories were divided into 70\% training and 30\% test set and the trajectories longer than 20 minutes were segmented into trips - sub trajectories - shorter than 20 minutes. 
Then, the authors applied the OPTICS clustering technique to show that the number of places where most people change their transportation modes has an upper bound limit\cite{zheng2010understanding}. 

In the spirit of the evaluation metrics applied in \cite{zheng2008understanding}, the \textit{accuracy by segment} ($A_s$), and \textit{accuracy by distance} ($A_d$), are proposed as two investigated evaluation methods. 
The former, $A_s$, represents the number of segments where the model correctly predicts their transportation modes, over the total number of segments. 
The second technique takes the length (distance) of trajectories into account for evaluation. This means each correctly predicted transportation mode has the trajectory segment distance as a coefficient. 
Therefore, the $A_d$ equals the length of trajectories correctly predicted over the length of total trajectories.
The best result reported for $A_s$ is \textbf{65.3} and the best result reported for $A_d$ is 72.8\% (Table 3  in \cite{zheng2008understanding}). 
Methods of handling noise data were not investigated in this research \cite{zheng2008understanding,Zheng2008}.


Endo et al. (2016) explored the transportation modes estimation while they believed hand-crafted features could not estimate transportation modes. 
They claimed that uncertainty in the data such as noise and diversity of human behavior are two main factors that mean hand-crafted features "cannot always work" \cite[P. 1]{endo2016deep}. 
This research generates a trajectory image from the segmented raw trajectory and extracts features using a deep neural network. 
This supervised approach is evaluated using the Geolife Dataset and the \textit{Kanto Trajectories}.

While the trajectory image generation process that proposed in \cite{endo2016deep} is a creative example of converting trajectory data to raster data, the spatial aspect of geographical trajectory data (e.g., happening in different countries) is neglected. 
Trajectories happening in a 3D spherical space and transforming them to 2D needs transformation. 
There are different projections that transform 3D to 2D such as azimuthal, conic, and cylindrical projection that could be used in image generation.

Some basic features are introduced by Zheng et al. like speed mean, acceleration mean and bearing mean, other advanced features (e.g., Heading change rate(HCR), Stop rate(SR), and velocity change rate(VCR)
), and the features generated by the deep neural network are generated and evaluated using a five fold cross validation approach under the conditions explained as follows. 
The training and test dataset are divided with 80\% training set and 20\% test set so that each user can appear only in either the training or test set. 
The segmentation approach used the annotation provided by users for each trajectory and users with less than ten annotations are removed from the data during the data preparation process.

There is a latent assumption in segmentation method of \cite{endo2016deep}, which is invalid, that the test set labels are available for doing segmentation. 
The targets of their research were seven transportation modes (walking, bus, car, bike, taxi, subway, and train) that are predicted with 67.9\% accuracy in the best set of generated features \cite{endo2016deep}. 
This method of calculating accuracy is equivalent to the accuracy by segment which is introduced by Zheng et al. (2008). 
Although the accuracy reported in \cite{endo2016deep} is lower than the accuracy achieved by Zheng et al. (2008)\cite{zheng2008understanding}, the comparison of these two approaches is not fair since they have different settings for splitting the data. 
Moreover, one big advantage of this model is being robust to the noise.


Another deep learning approach is investigated by Dabiri et al. (2018) in \cite{dabiri2018inferring}. 
In this approach, different configurations of a Convolutional Neural Network (CNN) are investigated. The inputs of the CNN are \textit{point features} like speed, acceleration, jerk, and bearing generated for each trajectory. 
The output of the network is set to the label of the corresponding trajectory. 
Trajectories are segmented using a pre-defined time interval threshold so that if two contiguous points have a time difference more than a pre-defined threshold, they are considered as a segment. 
Since they fed data into a CNN network with a fixed length of input, the fixed length is applied to subdivide the segments. 

The pre-processing task in \cite{dabiri2018inferring} starts with removing GPS points with time-stamp greater than their next point in the original order of data. 
They provided information about the maximum speed and acceleration in table 2 of \cite{dabiri2018inferring} and discarded the trajectories with unrealistic speed and acceleration, using the ground truth noise removal approach. 
This process precedes splitting the data for training and test. 
For the second step of their pre-processing task, they applied the Savitzky-Golay filter. 
Their CNN model used four sets of Conv-Conv-MaxPool-Dropout followed by three fully connected-dropout layers.

Dabiri et al.\cite{dabiri2018inferring} applied categorical cross entropy as loss function and Adam optimizer. 
The settings for their experiment is as follows. 
First, they merged car and taxi sub-trajectories and called them driving. 
Moreover, they merged the rail-based classes such as train, and subway as train. Therefore, their target classes are walk, bike, bus, driving and train- five classes. 
They randomly selected 80\% of the segments as their training set and the rest as the test set \cite{dabiri2018inferring}. 
Their best accuracy for the aforementioned targets is reported 84.8\% for the test set and Figure 3 of their paper shows an accuracy of more than 95\% on their training set. 
They explained that their model did not over-fit because the training accuracy is stable by performing more iterations \cite{dabiri2018inferring}.

Liu et al. (2017) proposed a Bidirectional Long Short Term Memory(LSTM) model to predict transportation modes\cite{liu2017end}. 
They fed their model with two inputs including mapping for time intervals and geo-location of trajectories. 
They applied a set of bi-LSTM for processing latitude and longitude changes and an embeddings layer for time intervals. 
They passed the results of two sub networks to a merged fully connected layer and getting the maximum probability using a softmax layer. 
They divided the training(80\%) and test(20\%) set with the mixed set of users, random cross validation. 
They compared their model with random forest and SVM using features introduced in \cite{zheng2008understanding}. 
To evaluate their model, they used the area under the curve (AUC) metric and reported 94.6\% AUC for their best model\cite{liu2017end}.

Xiao et al. (2017) studied the transportation modes prediction problem using tree-based ensemble classifiers. 
They believed other research studies in this field are not applicable all the time because they applied "multiple sensors or matching GIS information"\cite[P.1\&3]{xiao2017identifying}. 
Tree-based models such as random forest, gradient boosting decision tree and XGBoost were applied to achieve the best performance. 
They generated two categories of features called global features - 72 features including but not limited to mean, standard deviation, and mode - and local features - 39 features including but not limited to the number of decomposition class changes.

Xiao et al. \cite{xiao2017identifying} focused on classifying six transportation modes including walking, bus and taxi, bike, car, subway, and train\cite{xiao2017identifying}. 
They applied two steps of pre-processing. 
First they removed all the duplicate data points. 
Then, they removed abnormal trajectories using average speed, the ground truth noise removal approach. 
For example they discarded all the trajectories of walking whose average speed exceeds 10m/s. 
In order to evaluate their model, they splitted extracted trajectories into the training (70\%) and test (30\%) set randomly, random cross validation. 
They measured the accuracy, precision, recall, F-score and ROC since the target was unbalanced using the five fold cross validation technique. 
They reported XGBoost with 90.77\% accueacy as the best model in their research\cite{xiao2017identifying}.

 In order to predict the transportation modes, Zhu et al. (2018) \cite{zhu2018transportation} introduced a new segmentation method built on the idea of detecting rapid and sustained change in direction or speed. 
 They focused on seven transportation modes including walking, car, bus, bike, train, and plane. 
 The timeslice type, and acceleration rate were introduced as two new features. 
 Data was divided randomly to the training (70\%) and test(30\%) set\cite{zhu2018transportation}. 
 Moreover, a threshold for speed and acceleration was applied to clean data, i.e. the ground truth noise removal approach. 
 Accuracy by distance was used to evaluate their model and they report 89.31\% accuracy. 
 Using the plane (Airplane) transportation mode that clearly have different speed features and reporting 12,185 KM travel for the plane transportation mode are the two salient points that need to be considered.

The computational resources need for training of all the research studies above can be categorized into two groups. 
Some research studies applied conventional machine learning algorithms which required low level of computational resources.
For example, Zhu et al. (2018)\cite{zhu2018transportation}, Xiao et al. (2017)\cite{xiao2017identifying}, and Zheng et al. (2010)\cite{zheng2010understanding} applied models that can be trained on a CPU structure. 
On the other hand, some studies using deep learning methods need more computational resources. 
For instance, Dabiri et al. (2018)\cite{dabiri2018inferring}, Liu et al. (2017) \cite{liu2017end}, and Endo et al. (2016)\cite{endo2016deep} applied deep neural network models such as convolutional neural network (CNN) and bidirectional LSTM which require a GPU processing structure. 

Another aspect of the studies above is the features applied in their models. 
Representation learning, hand craft feature engineering and the combination of both are three categories in regards to applied features. 
Endo et al. (2016) \cite{endo2016deep}, Dabiri et al.(2018)\cite{dabiri2018inferring}, and Liu et al.(2017)\cite{liu2017end} are examples of using representation learning. 
They trained their model to extract a representation from the training set. 
Zhu et al.(2018) \cite{zhu2018transportation}, Xiao et al.(2016)\cite{xiao2017identifying}, and Zheng et al.(2010)\cite{zheng2010understanding} applied hand craft features. 
Endo et al. (2016)\cite{endo2016deep} took the advantage of both representation learning and hand craft features.

Feature engineering is a very important part of building a learning algorithm. Some of the algorithms extract features using representation learning methods; On the other hand, some studies select a subset of features from the hand craft features. Both methods have advantages such as learning faster, require less storage space, improving performance of learning and building more generalized models\cite{li2017feature}. However, there is two main difference. First, extracting features create new features where selecting features use a subset of available features. Second, selecting features constructs more readable and interpretable models\cite{li2017feature}.

Evaluation methods in these studies are mostly accuracy of models. 
However, the accuracy is calculated using different methods including random cross validation, cross validation with dividing users, cross validation with mix users and simple division of the training and test set without cross validation. 
The latter is a weak method that is used only in Zhu et al.(2018). 
The random cross validation or the conventional cross validation was applied in Xiao et al.(2017), Liu et al.(2017), and Dabiri et al.(2018). 
Zheng et al.(2010) mixed the training and test set according to users so that 70\% of trajectories of a user goes to the training set and the rest goes to test set. 
Only Endo et al.(2016) performed the cross validation by dividing users between the training and test set. 
Because trajectory data is a kind of data with spatio-temporal dimensions and possibility of having users in the same semantic hierarchical structure such as students, worker, visitors, and teachers, the conventional cross validation method might provide optimistic results as studied in \cite{roberts2017cross}.

Smoothing, ground truth are no removal are three different approaches to handle noise in the reviewed studies. Zheng et al.(2010) and Endo et al.(2016) did not mention anything related to noise removal methods. Xiao et al.(2017), Zhu et al.(2018), and Dabiri et al.(2018). applied a ground truth knowledge about speed to remove abnormal trajectories. 
Dabiri et al (2018) applied a smoothing filter to remove the GPS error in their pre-processing step.



\begin{table}[ht]
\centering
\caption{Summary of the related work}
\label{tab:summary}
\begin{tabular}{|l|c|l|l|l|l|l|c|c|}
\hline
\multicolumn{1}{|c|}{\textbf{\rotatebox[origin=c]{90}{Study}}} & \textbf{\rotatebox[origin=c]{90}{Computational   }\rotatebox[origin=c]{90}{requirement }} & \multicolumn{1}{c|}{\textbf{\rotatebox[origin=c]{90}{Features }}} & \multicolumn{1}{c|}{\textbf{\rotatebox[origin=c]{90}{Noise handling }}} & \multicolumn{1}{c|}{\textbf{\rotatebox[origin=c]{90}{Evaluation }}} & \multicolumn{1}{c|}{\textbf{\rotatebox[origin=c]{90}{Metric }}} & \multicolumn{1}{c|}{\textbf{labels}} & \textbf{\rotatebox[origin=c]{90}{Reported }\rotatebox[origin=c]{90}{score }} & \textbf{\rotatebox[origin=c]{90}{\# of segments }} \\
 & &  &  &  &  &  & \multicolumn{1}{l|}{} & \multicolumn{1}{l|}{} \\ \hline
\textbf{\cite{endo2016deep}} & GPU & \begin{tabular}[c]{@{}l@{}}RL$^{\mathrm{1}}$ \\  HC$^{\mathrm{2}}$\end{tabular} & N$^{\mathrm{3}}$ &UCV$^{\mathrm{6}}$ &$A_s$ & \begin{tabular}[c]{@{}l@{}}7 classes:\\ walking, bus,\\ car, bike, taxi, \\ subway, and train\end{tabular} & 67.9\% & 9043 \\ \hline
\textbf{\cite{dabiri2018inferring}} & GPU  & RL$^{\mathrm{1}}$ & \begin{tabular}[c]{@{}l@{}}S$^{\mathrm{4}}$ \\ G$^{\mathrm{5}}$\end{tabular} & RCV$^{\mathrm{7}}$&$A_s$ & \begin{tabular}[c]{@{}l@{}}5 classes:\\ walk, bike, bus, \\ driving, and train\end{tabular} & 84.8\% & 32444 \\ \hline
\textbf{\cite{xiao2017identifying}} & CPU  & HC$^{\mathrm{2}}$ & G$^{\mathrm{5}}$ & RCV$^{\mathrm{7}}$ &$A_s$ & \begin{tabular}[c]{@{}l@{}}6 classes:\\walk, bus\&taxi, \\ bike, car, \\ subway, train\end{tabular} & 90.77\% & 7985 \\ \hline
\textbf{\cite{liu2017end}} & GPU  & RL$^{\mathrm{1}}$ & N$^{\mathrm{3}}$ & NoCV$^{\mathrm{8}}$ & AUC & N/M$^{\mathrm{11}}$ & 94.6\% & 17621 \\ \hline
\textbf{\cite{zheng2010understanding}} & CPU  & HC$^{\mathrm{2}}$ & N$^{\mathrm{3}}$ &NoCV$^{\mathrm{9}}$ & \begin{tabular}[c]{@{}c@{}}$A_s$\\ $A_d$\end{tabular} & \begin{tabular}[c]{@{}l@{}}4 classes:\\ walk, driving, \\ bus, bike\end{tabular} & \begin{tabular}[c]{@{}c@{}}65.3\%\\ 76.8\%\end{tabular} & 7112 \\ \hline
\textbf{\cite{zhu2018transportation}} & CPU  & HC$^{\mathrm{2}}$ & S$^{\mathrm{4}}$ &NoCV$^{\mathrm{10}}$ & $A_d$ & \begin{tabular}[c]{@{}l@{}}6 classes:\\ walk, car, bus\\ bike, train, plane\end{tabular} & 89.31\% & N/M$^{\mathrm{11}}$ \\ \hline
\multicolumn{6}{l}{$^{\mathrm{1}}$Representation Learning}\\
\multicolumn{6}{l}{$^{\mathrm{2}}$Hand Craft features}\\
\multicolumn{6}{l}{$^{\mathrm{3}}$No noise removal}\\
\multicolumn{6}{l}{$^{\mathrm{4}}$Smoothing}\\
\multicolumn{6}{l}{$^{\mathrm{5}}$The ground truth}\\
\multicolumn{8}{l}{$^{\mathrm{6}}$User oriented five folds cross validation}\\
\multicolumn{8}{l}{$^{\mathrm{7}}$Random five folds cross validation}\\
\multicolumn{8}{l}{$^{\mathrm{8}}$
No cross validation Train80\% and Test(20\%)}\\
\multicolumn{8}{l}{$^{\mathrm{9}}$
No cross validation and each user 70\% for training and 30\% test}\\
\multicolumn{8}{l}{$^{\mathrm{10}}$
Divide data to an offline training and  online prediction}\\
\multicolumn{8}{l}{$^{\mathrm{11}}$
Not mentioned}\\
\multicolumn{8}{l}{ $A_s$: Accuracy by segment,$A_d$: Accuracy by distance}\\
\end{tabular}
\end{table}
\chapter{Preliminaries}
\label{sec:Preliminaries}

\section{Notation and Definitions}

A \emph{trajectory point}, $l_i \in L$, is defined in notation \ref{not:1}, where $x_i$ is longitude varies from 0$^{\circ}$ to $\pm 180^{\circ}$, $y_i$ is latitude varies from 0$^{\circ}$ to $\pm 90^{\circ}$, and $t_i$ ($t_i < t_{i+1}$) is the capturing time of the moving object and $L$ is the set of all trajectory points. 
\begin{equation}
l_i=(x_i,y_i,t_i)
\label{not:1}
\end{equation}
A trajectory point can be assigned by some features that describe different attributes of the moving object with a specific time-stamp and location. 
The time-stamp and location are two dimensions that make trajectory point  \emph{spatio-temporal} (ST) data. 
This type of data has two important properties: (i) \emph{auto-correlation} and (ii) \emph{heterogeneity} \cite{STDM2017}.

\emph{Auto-correlation} means that two trajectory points at a nearby location and time are highly correlated. 
The first law of geography is ``everything is related to everything else, but near things are more related than distant things."\cite{tobler1970computer}.
Therefore cross-validation may become invalid since the randomly generated training and test sets are correlated \cite{STDM2017}. 
Extracting trajectory features from raw trajectories makes sub trajectories at nearby time and location correlated as well.
Therefore, the cross-validation of trajectory samples may become invalid since training and test sets are still correlated. 
Applying some conventional machine learning methods such as filter feature selection methods that utilized chi2 is also invalid because of the violation of the chi2's assumption that the samples are independent.

\emph{Heterogeneity} means that a trajectory point might come from different populations; while, two samples in conventional data mining come from the same population. 
For example, data collected from a GPS device of a car in a rush hour comes from different populations than the data collected in normal traffic.


A \emph{raw trajectory}, or simply a trajectory, is a sequence of trajectory points captured through time. 
\begin{equation}
\tau=(l_i,l_{i+1},..,l_n), l_j \in L, i \leq n
\label{not:2}
\end{equation}

A \emph{trajectory label}, or label, $o_i \in O$ is an annotation of a trajectory that is a categorical variable of transportation modes and $O$ is the set of labels or transportation modes.
Notation \eqref{not:3} shows an example set of trajectory labels in this work.
\begin{equation}
o_i \in \{Bus, Walk, Train, Bike\}
\label{not:3}
\end{equation}

A \emph{sub-trajectory} is a consecutive sub-sequence of a raw trajectory generated by splitting the raw trajectory into two or more sub-trajectories. 
For example, if we have one split point, $k$, and $\tau_1$ is a raw trajectory then $s_1=(l_i,l_{i+1},...,l_{k})$ and $s_2=(l_{k+1},l_{k+2},...,l_n)$ are two sub trajectories generated by $\tau_1$. 
The process of generating sub trajectories from a raw trajectory is called \emph{segmentation}.

Several segmentation methods have been proposed in recent years including but not limited to temporal based \cite{stenneth2011transportation,dabiri2018inferring}, cost function based \cite{soares2015grasp,grasp-semts2018}, and semantic based methods \cite{spaccapietra2008conceptual}.
The focus of this research is to compare the transportation modes prediction methods so the topic of trajectory segmentation is not explored in this work. 
Therefore, we used a daily segment of raw trajectories and then segmented the data using the transportation modes annotations to partition the data. This approach is also used in  \cite{dabiri2018inferring,endo2016deep}. 
The assumption that the transportation modes are available for test set segmentation is invalid since we are going to predict them by our model; 
However, we need to prepare a controlled environment similar to \cite{dabiri2018inferring,endo2016deep} to study the performance of the transportation modes prediction.

A \emph{point feature} is a measured value $F_p$, assigned to each trajectory points of a sub trajectory $S$. 
The notation \ref{not:4} shows the feature $F_p$ for sub trajectory $S$. 
For example, speed can be a point feature since we can calculate the speed of a moving object for each trajectory point. 
Since we need two trajectory points to calculate speed, we assume the speed of the first trajectory point is equal to the speed of the second trajectory point.

\begin{equation}
Fˆp=(f_i,f_{i+1},..,f_n)
\label{not:4}
\end{equation}
\label{sec:trajfeatures}
A \emph{trajectory feature} is a measured value $F_t$, assigned to a sub trajectory, $S$. 
The notation \ref{not:5} shows the feature $F_t$ for sub trajectory $S$. 
For example, the speed mean can be a trajectory feature since we can calculate the speed mean of a moving object for a sub trajectory. 

The $F_t^p$ is the notation for all trajectory features that generated using point feature $p$. For example, $F_t^{speed}$ represents all the trajectory features derived from $speed$ point feature. Moreover, $F_{mean}^{speed}$ represents the mean of the trajectory features derived from the $speed$ point feature.

\begin{equation}
F_t= \frac{\Sigma f_k}{n}
\label{not:5}
\end{equation}

\section{GeoLife Dataset}
\label{sec:GeoLifeDataset}

\begin{figure}[htb]
\caption{Distribution of $F_{mean}^{speed}$ for walk, bike, bus, and car}
\centering
\includegraphics[width=0.8\textwidth]{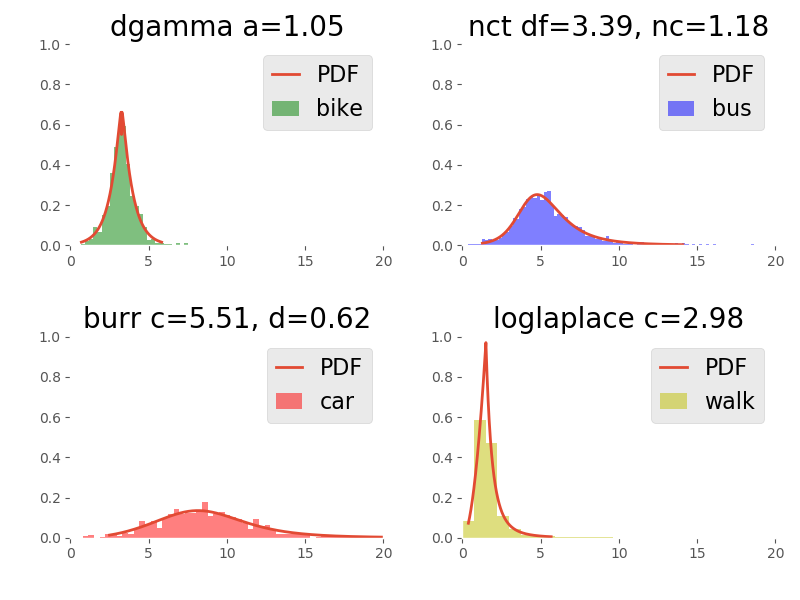}
\label{fig:dist}
\end{figure}

In this work, we focus on the GeoLife dataset \cite{geolife-gps-trajectory-dataset-user-guide}.
This dataset has 5,504,363 records collected by 69 users, and is labeled with eleven transportation modes: taxi (4.41\%); car (9.40\%); train (10.19\%); subway (5.68\%); walk (29.35\%); airplane (0.16\%); boat (0.06\%); bike (17.34\%); run (0.03\%); motorcycle (0.006\%); and bus (23.33\%).
Figure \ref{fig:dist} shows the distribution of $F_{mean}^{speed}$ for four transportation modes including walking, bus, bike, and car. 
This chart is generated by searching more than 80 statistical distributions to find the best parameters to fit the $F_{mean}^{speed}$ data. 
The \emph{dgamma} distribution with a=1.05, loc=3.27, and scale=0.64 shows the behaviour of \textit{bike} $F_{mean}^{speed}$. 
The \emph{nct} with df=3.39, nc=1.18, loc=3.42, and scale 1.37 represents \textit{bus} $F_{mean}^{speed}$ behavior. 
The \emph{burr} with c=5.51, d=0.62, loc=-0.03 and scale=9.81 represents \textit{car} $F_{mean}^{speed}$ behaviour and the \emph{logloplace} with c=2.98, loc=-0.03 and scale=1.53 shows a \textit{walking} $F_{mean}^{speed}$ behaviour. 
The probability distribution function (PDF) was applied to calculate the minimum and maximum threshold as the ground truth as is applied in \cite{zhu2018transportation,xiao2017identifying,dabiri2018inferring}. 
Table \ref{tab:bounds} shows the details of upper and lower bounds for different transportation modes.

\begin{table}[htb]
\centering
\caption{Upper and lower bounds for different transportation modes according to the best fit distribution.}
\label{tab:bounds}
\begin{tabular}{|l|c|c|c|c|c|c|}
\hline
\multicolumn{1}{|c|}{$F_{mean}^{speed}$} & \textbf{car} & \textbf{bus} & \textbf{bike} & \textbf{taxi} & \textbf{train} & \textbf{walk} \\ \hline
\textbf{lower bound}             & 2.502        & 1.278        & 0.703         & 1.923         & 1.953          & 0.379         \\ \hline
\textbf{upper bound}             & 20.629       & 14.084       & 5.832         & 17.214        & 52.957         & 5.673         \\ \hline
\end{tabular}
\end{table}

\section{Veracity of data}
Handling abnormality in the data, missing values, and noise are referred to as veracity. 
They address uncertainties, truthfulness and trustworthiness of data. 
Two main sources of uncertainty of data in the trajectory mining tasks are device error and human error which are reviewed in the following sections.

\label{sec:UncertaintyofData}
\subsection{Device error}
GPS records captured by a device have always some kind of inaccuracy. 
This inaccuracy can be categorized in two major groups, systematic errors and random errors \cite{jun2006smoothing}. 
The \textit{systematic error} occurs when the recording device cannot find enough satellites to provide precise data. 
In this case, the position dilution of precision (PDOP) is high. 
The \textit{random error} can happen because of atmospheric and ionospheric effects. 
The \textit{systematic errors} are easier to detect while the \textit{random errors} are more difficult. 
For cleaning the GPS data, there are different filtering methods including but not limited to hampel filter\cite{hampel1974influence}, Kalman filter \cite{jun2006smoothing,morrison2014introduction}, and Savitzky-Golay filter\cite{Savitzky-GolayFilter}. 
Although an extension of a Kalman filter provides the best results for removing noise \cite{jun2006smoothing}, it consumes a lot of computational power since it has an iterative nature (expectation - maximization). 
Moreover, most GPS devices perform a kind of embedded Kalman filter as their pre-processing before capturing data. 
The Savitzky-Golay filter fits a polynomial function to a fixed window and the hampel filter works based on the median of a fixed window and is the simplest method. 
For example, Figure \ref{fig:noise_1} shows a trajectory annotated as \textit{bike}. 
In this figure, the size of marker is relative to the speed of the moving object and the marker pointer direction is relative to the bearing, the angle between moving object direction vector and its direction vector pointed to north, of the moving object. 
In Figure \ref{fig:noise_1}, we show some GPS errors with blue and yellow color.

\subsection{Human error}
The data annotation process has been done after each tracking as \cite{zheng2008understanding} explained in the Geolife dataset documentation.
As humans we are all fallible.
Therefore, it is possible that some users forget to annotate the trajectory when they switch from one transportation mode to another. 
Although this is an assumption and it can not be proved, some trajectory samples, some circumstantial evidence, show a continuous behaviour change that is suspected to be this type of error. 

\begin{figure}[htb]
\caption{Some uncertainty of GPS Data}
\centering
\includegraphics[width=0.8\textwidth]{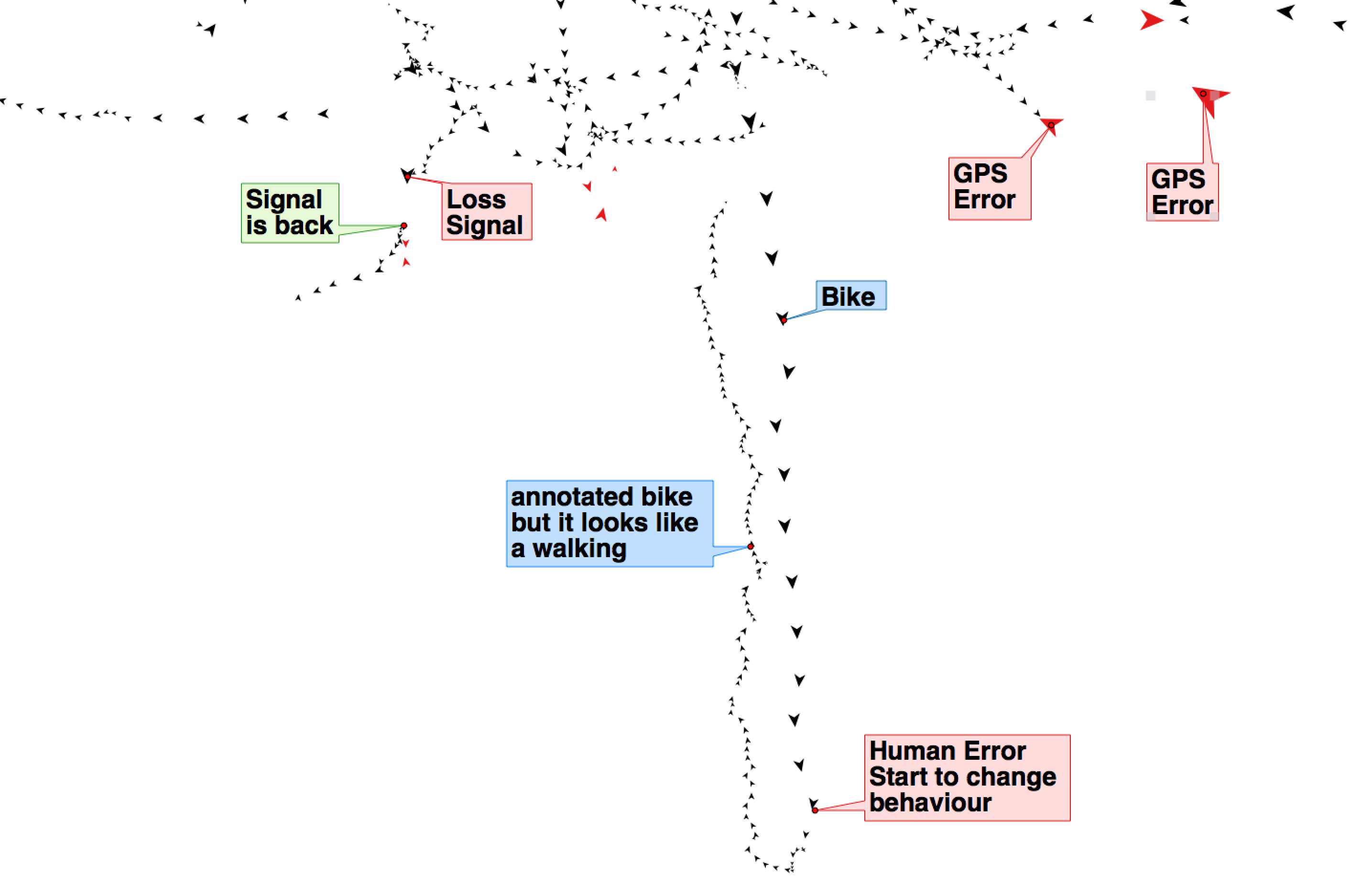}
\label{fig:noise_1}
\end{figure}

Figure \ref{fig:noise_1} shows a trajectory labeled as bike in the Geolife dataset generated by Q-GIS version 2.8. 
The marker size in this figure is set to \textit{speed} and the orientation set to \textit{bearing}. 
The red markers show GPS errors. 
The changes in the speed pattern (changes in the size of marker) might be a representation of human error.
For instance, in Figure \ref{fig:noise_1}, we observe that there is a behaviour change that can possibly be a transition from bike to walk. 
We cannot prove this behaviour was not bike because the user might be walking with their bike!


We assume the Bayes error is the minimum possible error and human error is near to the Bayes error.
Avoidable bias is defined as the difference between the training error and the human error\cite{andNGavoidablebias}. 
Achieving the performance near to the human performance in each task is the main objective of a research. The recent advancements in deep learning leads to achieve some performance level even more than the performance of doing the task by human because of using huge samples and scrutinizing the data to fine clean it. 
However, ``we cannot do better than Bayes error unless we are overfitting" \cite{andNGavoidablebias}.
Having noise in GPS data and human error suggest the idea that the avoidable bias is not equal to zero. 
This ground truth was our base to include a research results in our related work or exclude it.

\section{The framework}
\label{sec:model}

\begin{figure}[ht]
\caption{The steps of the applied framework to predict transportation modes}
\centering
\includegraphics[width=0.8\textwidth]{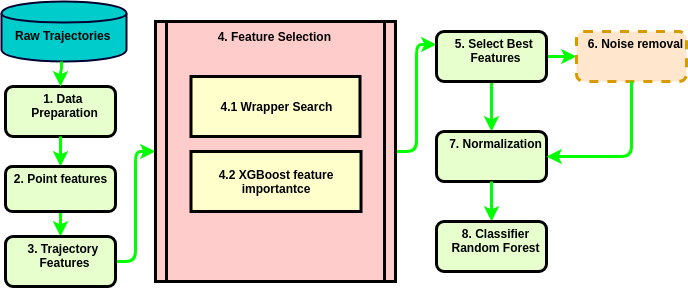}
\label{fig:model}
\end{figure}

In this section, the sequence of steps of the framework with eight steps are explained (Figure \ref{fig:model}).

The first step groups the trajectory points by \textit{user id}, \textit{day} and \textit{transportation modes} to create sub trajectories (segmentation). 
Sub trajectories with less than ten trajectory points were discarded to avoid generating low quality trajectories. 

Point features including speed, acceleration, bearing, jerk, bearing rate, and the rate of the bearing rate were generated in step two.
The features speed, acceleration, and bearing were first introduced in \cite{zheng2008understanding} and, jerk was proposed in \cite{dabiri2018inferring}. 
The very first point feature that we generated is duration. 
This is the time difference between two trajectory points. 
This feature gives us very important information including some of the segmentation position points, loss signal points, and is useful in calculating point features such as speed, and acceleration.

The distance between two trajectory points is calculated using haversine distance\cite{haversine}. 
Since we work with a lot of trajectory points, we need an efficient way to calculate the haversine in each trajectory. 
Therefore, the haversine code was rewritten in a vectorized manner in Python programming language which is much faster than the original code published in \cite{haversine}. 
This implementation is available at \cite{trajLib}.

Having duration and distance as two point features, we calculate speed, acceleration and jerk using Equation \ref{eq:speed},\ref{eq:acc}, and \ref{eq:jerk}, respectively.

\begin{equation}
\label{eq:speed}
S_{i}=\frac{Distance_{i}}{Duration_{i}}
\end{equation}

\begin{equation}
\label{eq:acc}
A_{i+1}=\frac{(S_{i+1}-S_{i})}{\Delta t}
\end{equation}

\begin{equation}
\label{eq:jerk}
J_{i+1}=\frac{(A_{i+1}-A_{i})}{\Delta t}
\end{equation}

For calculating the bearing point feature, we rewrite a vectorized version of the compass bearing code available in \cite{compassbearing}. 
The updated code is available in \cite{trajLib}.

Two new features are introduced in \cite{etemad2018predicting}, named bearing rate, and the rate of the bearing rate. 
Applying equation \ref{eq:5}, we computed the bearing rate. 
$B_i$ and $B_{i+1}$ are the bearing point feature values in points $i$ and $i+1$.
$\Delta t$ is the time difference\cite{etemad2018predicting}. 

\begin{equation}
\label{eq:5}
B_{rate(i+1)}=\frac{(B_{i+1}-B_{i})}{\Delta t}
\end{equation}

The rate of the bearing rate point feature is computed using Equation \ref{eq:6}.

\begin{equation}
\label{eq:6}
Br_{rate(i+1)} =\frac{(B_{rate(i+1)}-B_{rate(i)})}{\Delta t}
\end{equation}

After calculating the point features for each trajectory, 
the trajectory features were extracted in step three, using our trajectory library \cite{trajLib}. 
Trajectory features are divided into two different types including global trajectory features and local trajectory features.

Global features, like the Minimum, Maximum, Mean, Median, and Standard Deviation, summarize information about the whole trajectory and local trajectory features, like percentiles (e.g., 10, 25, 50, 75, and 90), describe a behaviour related to part of a trajectory. 

The local trajectory features extracted in this work were the percentiles of every point feature.
Five different global trajectory features were used in the models tested in this work. 
In summary, we compute 70 trajectory features (i.e., 10 statistical measures including five global and five local features calculated for 7 point features) for each transportation mode sample. 
In Step 4, two feature selection approaches were performed,  wrapper search and information retrieval feature importance. According the best accuracy results for cross validation, a subset of top 20 features were selected in step 5.

In step 6, the framework deals with noise in the data optionally. This means that we ran the experiments with and without this step.
We explain different methods of noise removal in section \ref{sec:Noiseremovalmethods}.
Finally, we normalized the features (step 7) using the Min-Max normalization method, since this method preserves the relationship between the values to transform features to the same range and improves the quality of the classification process \cite{han2011data}. 

\section{Noise removal methods}
\label{sec:Noiseremovalmethods}
In the literature, we found three major strategies of noise removal, smoothing\cite{dabiri2018inferring}, 
the ground truth\cite{dabiri2018inferring,xiao2017identifying,zhu2018transportation}, and clustering techniques\cite{etemad2018predicting}. 
In this section we explain them and in the section \ref{sec:experiments} we investigate their effects of classification.


Smoothing is a method that can handle GPS errors explained in section \ref{sec:UncertaintyofData}. 
These methods try to fix GPS measurement for latitude and longitude. 
Jun et al.\cite{jun2006smoothing} show that the extended Kalman filter has the best performance for the smoothing task. 
This method is in the family of expectation maximization algorithm \cite{jun2006smoothing}. 
Therefore, the algorithm needs to iterate to converge for each window. 
Moreover, most GPS devices have a simplified version of it embedded in their firmware. 
The simplest method is hampel filter\cite{hampel1974influence}. 
This filter gets the median of a fixed window and adjusts the points sitting outside of a threshold, usually relative to standard deviation of the fixed window. 
This type of noise removal can be done as a pre-processing step and can be done before generating point features.


The ground truth method lays on the idea that speed or acceleration of a trajectory has an upper and a lower bound. 
We can see this technique in some research studies as a pre-processing step before dividing the training and test set\cite{xiao2017identifying,dabiri2018inferring}. 
Although this is a widely used method of pre-processing, applying this technique in classification may not be the best approach because after dividing the dataset into the training and test, the goal of classification is to predict the labels for the test set. Therefore, we must assume we do not have access to the labels for the test set. 

The clustering method passes one variable, such as $F_{mean}^{speed}$, to a clustering algorithm like DBSCAN. 
Then the data is clustered and the outliers are found and removed. 
The problem with this method is that selecting the appropriate hyper-parameters such as epsilon value in DBSCAN.

\section{Feature selection}
Feature selection methods can be categorized in three general groups: filter models, wrapper models, and embedded models\cite{fs_guyon2003introduction}. 
Filter model methods are independent of the learning algorithm. 
They select features based on the nature of data regardless of the learning algorithm\cite{li2017feature}. 
On the other hand, wrapper methods are based on a kind of search, such as sequential, best first, or branch and bound, to find the best subset that gives the highest score on a selected learning algorithm\cite{li2017feature}. 
The embedded methods apply both filter and wrapper\cite{li2017feature}.

Feature selection methods can be grouped based on the type of data as well. 
The feature selection methods that use the assumption of (Independent and identically distributed)i.i.d. are conventional feature selection methods\cite{li2017feature} such as \cite{he2005laplacian,zhao2007spectral,li2012unsupervised,nie2008trace,peng2005minimum,peng2005feature,liu1995chi2}.
They are not designed to handle heterogeneous or auto-correlated data. 
Some feature selection methods have been introduced to handle heterogeneous data and stream data that most of them working on graph structure such as \cite{gu2011towards,tang2012unsupervised,li2017toward}.
There is no specific feature selection method that is designed for trajectory data.

Conventional feature selection methods are categorized in four groups: similarity based methods like \cite{he2005laplacian}, Information theoretical methods like \cite{peng2005feature}, sparse learning methods such as \cite{li2012unsupervised}, and statistical based methods like \cite{liu1995chi2}.
In the reviewed literature, we did not find any feature selection algorithm specifically for Trajectory data that can handle auto-correlation, heterogeneity of data and moving object sensor behaviour. Therefore, perform two experiments using a wrapper method and a information theoretical method.

\chapter{Experiments}
\label{ch:experiments}
\label{sec:experiments}

In this section, we detail the five experiments performed in this work to investigate different aspects of our framework. 

The first experiment investigated among six classifiers, which classifier is the best. 
The experiment settings are set to a regular cross validation and to perform the transportation mode prediction task showed on \cite{dabiri2018inferring}. 

The second experiment is selecting the best features for transportation modes prediction task. 
The user oriented cross validation and random forest classifier were used for evaluation of transportation modes used in \cite{endo2016deep}.
The wrapper method implemented to search the best subset of our 70 features.
The information theoretical feature importance methods were used to select the best subset of our 70 features for the transportation modes prediction task. 

The third experiment is a comparison between \cite{endo2016deep} and our implementation. The user oriented cross validation, the top 20 best features, and random forest were applied to compare our work with \cite{endo2016deep}.

The forth experiment is another comparison between \cite{dabiri2018inferring} and our implementation.
The random cross validation on the top 20 features was applied to classify transportation modes used in \cite{dabiri2018inferring} using random forest classifier.

The last experiment is comparing two methods of cross validation, random cross validation and user oriented cross validation. In this experiment, the linear correlation between the training set and test set was compared. 

\section{Classifier selection}
The very first question in transportation modes prediction is which classifier performs better in this domain.
We applied XGBoost, SVM, Decision Tree, Random Forest, Neural Network, and Adaboost.
We applied SKLearn implementation for all the above algorithms with seed=10.
The parameters of the classifiers are adjusted as described in appendix \ref{sec:param} Table\ref{tab:param}. 

\begin{figure}[ht]
\caption{Among the trained classifiers random forest achieved the highest mean accuracy.}
\centering
\includegraphics[width=0.8\textwidth]{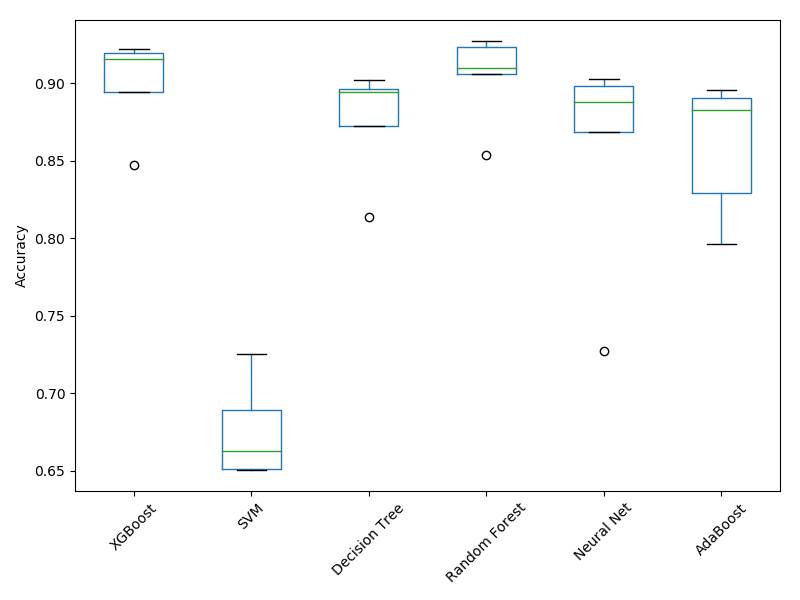}
\label{fig:clfscv}
\end{figure}

The dataset is filtered based on labels that have been applied in \cite{dabiri2018inferring} (e.g.,  walking, train, bus, bike, driving). 
No noise removal method was applied. 
The aforementioned classifiers were trained and the accuracy metric was calculated using random cross validation similar to \cite{liu2017end,xiao2017identifying,dabiri2018inferring}.
The results of cross validation, presented in Figure \ref{fig:clfscv}, show that the random forest performs better than other models($\mu_{accuracy} = 90.4\%$). 
The second best model was XGBoost ($\mu_{accuracy} = 90.00\%$).
A Wilcoxon Signed-Ranks Tests indicated that the random forest classifier results were not statistically significantly higher than the XGBoost classifier results.



The cross validation accuracy results show that the SVM algorithm is the weakest classifier between the investigated classifiers. 
The outliers in the results show that the cross validation method cannot find folds that are coming from the same distribution. 
That is, one of the folds has some samples that classifier cannot generalize behaviour of its samples from the rest of folds. 
The random forest not only get the highest mean accuracy but also it generates the least variance for the accuracy results. 
That means, applying the random forest classifier provides more reliability.

\section{Feature selection}
\subsection{Wrapper feature selection}
Wrapper feature selection method is a method that we select for implementation because it performs feature selection for a specific learning task which here is transportation modes prediction. 
In algorithm \ref{alg:1}, the details of this approach is provided. 
First, we define an empty set for selected features. Then, we search all the trajectory features one by one to find the best feature to append to the selected feature set.
The maximum accuracy score is the metric for selecting the best feature to append to selected features. Then we remove the selected feature from the set of features and repeat the search for union of selected features and next candidate feature in the feature set.

\begin{figure}[ht]
\caption{Accuracy of random forest classifier for incremental appending best features}
\centering
\includegraphics[width=\textwidth]{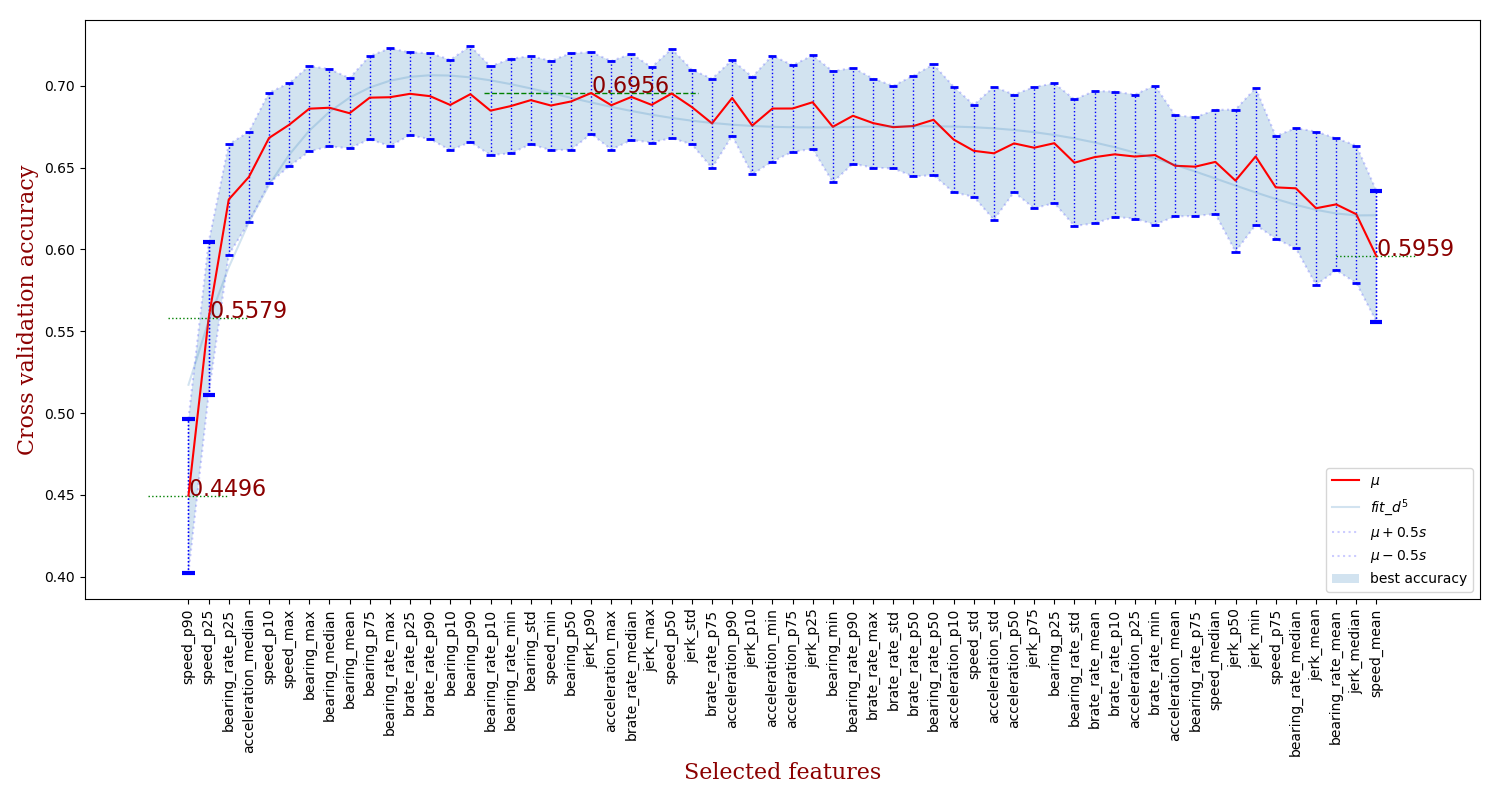}
\label{fig:fs}
\end{figure}

We select the labels applied in \cite{endo2016deep} and the same cross validation technique. 
The results are shown in Figure \ref{fig:fs}.
The results of this method suggest that the top 20 features gets the highest accuracy. Therefore we select this subset as the best subset for classification.
The best features ordered by their importance are $F_{p90}^{speed}$ (e.g. the percentile 90 of the speed as defined in section \ref{sec:trajfeatures}, notations \ref{not:4} and \ref{not:5}), $F_{p25}^{speed}$, $F_{p25}^{bearing\_rate}$, $F_{median}^{acceleration}$, $F_{p10}^{speed}$,$F_{max}^{speed}$, $F_{max}^{bearing}$, $F_{median}^{bearing}$, $F_{mean}^{bearing}$, $F_{p75}^{bearing}$, $F_{max}^{bearing\_rate}$, $F_{p25}^{brate\_rate}$, $F_{p90}^{brate\_rate}$, $F_{p10}^{bearing}$, $F_{p90}^{bearing}$, $F_{p10}^{bearing\_rate}$, $F_{min}^{bearing\_rate\_min}$, $F_{std}^{bearing}$, $F_{min}^{speed}$, and $F_{p50}^{bearing}$.

\begin{algorithm}
\label{alg:1}
\caption{Select the best features using wrapper search}
\begin{algorithmic} 
\STATE $features \longleftarrow {F_t^{speed}}\cup {F_t^{acceleration}}\cup {F_t^{bearing}}\cup {F_t^{jerk} }\cup {F_t^{brate}}\cup {F_t^{rbrate}}$
\STATE $selected\_features \longleftarrow \emptyset$
\STATE $cp\_features \longleftarrow all\_features$
\WHILE{$features \neq \emptyset$}
\STATE $dic\_cv\gets\emptyset$
\FORALL{$f\in\mathit{features}$}
\STATE $var\gets [f]\cup{selected\_features}$
\STATE $cv\_accuracy,cv\_mean\gets {learning\_task}(var)$
\STATE $dic\_cv[var]\gets cv\_accuracy,cv\_mean$
\ENDFOR
\STATE ${F_{max}}\gets {maximum}(dic\_cv)$
\STATE ${features}\gets {features}-{F_{max}}$
\STATE ${selected\_features}\gets {selected\_features}\cup(F_{max})$
\ENDWHILE
\RETURN{$selected\_features with highest score$}
\end{algorithmic}
\end{algorithm}

\subsection{Information theoretical feature selection}
Information theoretical feature selection is one of the methods widely used to select important features. 
XGBoost is a classifier that has embedded feature selection using information theoretical metrics.

\begin{figure}[ht]
\caption{Accuracy of random forest classifier for incremental appending features ranked by XGBoost feature importance}
\centering
\includegraphics[width=\textwidth]{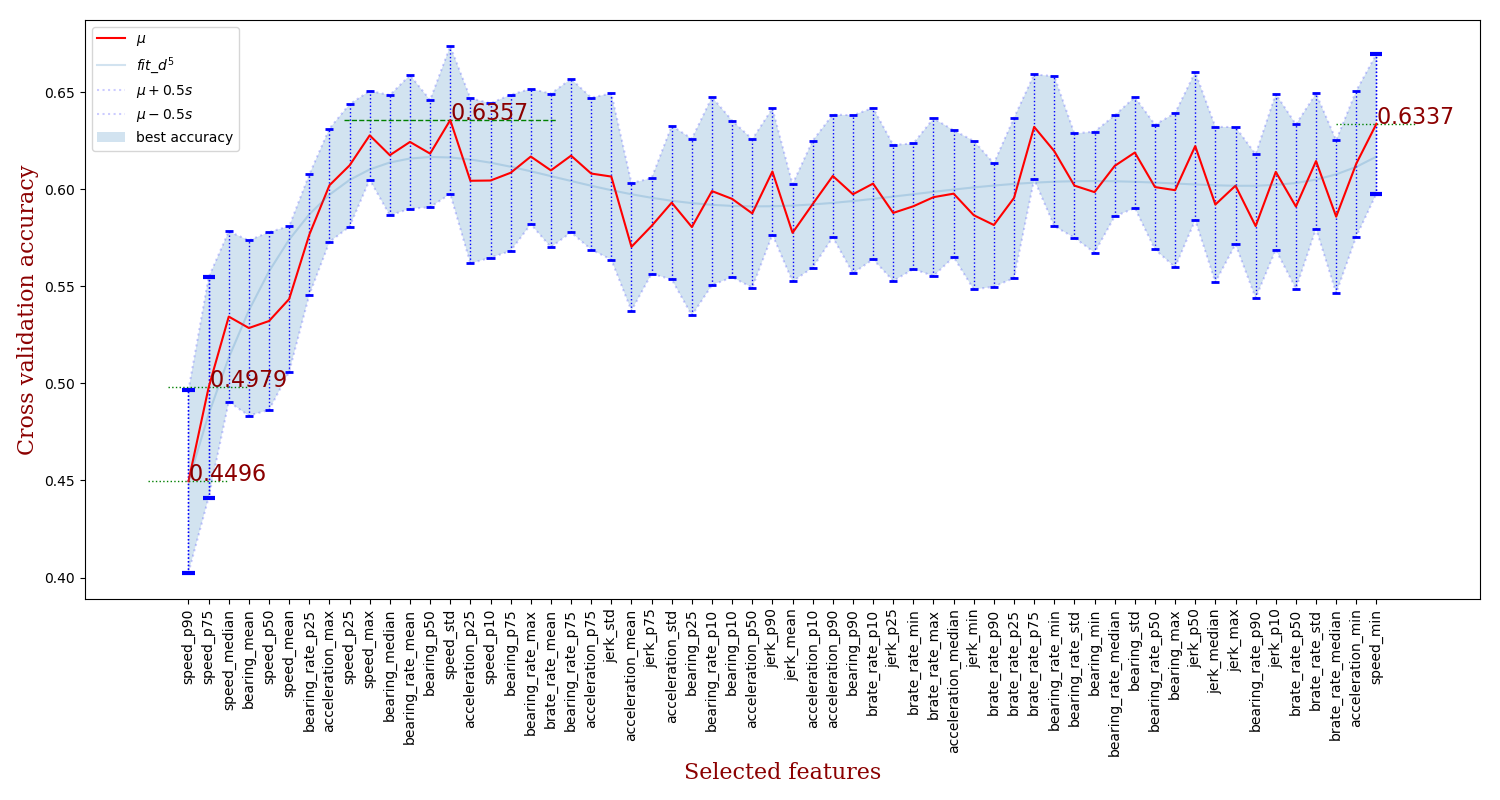}
\label{fig:fs2}
\end{figure}

In this experiment, we calculate the feature importance using XGBoost. 
Then, each feature is appended to the selected feature set and calculating the accuracy score for random forest classifier. In this experiment, the user oriented cross validation is used and the target labels are car, walking, bus, bike, taxi, and subway similar to \cite{endo2016deep}. 
Figure \ref{fig:fs2} shows the results of cross validation for appending features with respect to the importance rank suggested by the XGBoost.

\section{Comparison with Endo et al.[9]}

In this third experiment, we filtered transportation modes which have been used by Endo et al.\cite{endo2016deep} for evaluation. 
We divided the training and test dataset in a way that each user can appear only either in the training or test set. 
The top 20 features were selected to be used in this experiment which are $F_{p90}^{speed}$, $F_{p25}^{speed}$, $F_{p25}^{bearing\_rate}$, $F_{median}^{acceleration}$, $F_{p10}^{speed}$,$F_{max}^{speed}$, $F_{max}^{bearing}$, $F_{median}^{bearing}$, $F_{mean}^{bearing}$, $F_{p75}^{bearing}$, $F_{max}^{bearing\_rate}$, $F_{p25}^{brate\_rate}$, $F_{p90}^{brate\_rate}$, $F_{p10}^{bearing}$, $F_{p90}^{bearing}$, $F_{p10}^{bearing\_rate}$, $F_{min}^{bearing\_rate\_min}$, $F_{std}^{bearing}$, $F_{min}^{speed}$, and $F_{p50}^{bearing}$ (as defined in section \ref{sec:trajfeatures}, notations \ref{not:4} and \ref{not:5})).
Therefore, we approximately divided 80\% of the data as training and 20\% of the data as test. 
Endo et al.\cite{endo2016deep} reported accuracy per segment.
Thus, we compare our accuracy per segment results against their mean accuracy, 67.9\%.
In order to compare our cross validation results with the accuracy reported by Endo et al.\cite{endo2016deep}, we checked the normality of data and applied appropriate statistical test.
We cleaned data using DBSCAN method by removing outliers for one experiment and did not clean the data for another experiment.
The Shapiro-Wilks test(pvalue=0.0405), kolomogorov-Smirnov test(p-value=0.0035), and the number of samples suggest to use non-parametric tests.

Since the data does not come from normal distribution, we chose a non parametric test to compare the distribution of our observations with reported accuracy. 
A one sample wilcoxon rank sum test is the most appropriate statistical test to compare our data with the reported accuracy. 
A one sample Wilcoxon Signed-ranks test indicated that our accuracy results (69.50\%) is higher than Endo et al.\cite{endo2016deep}'s results (67.9\%), p=0.0431.

We repeated this experiment without using the noise removal step to understand whether there is any difference between our model and Endo's when our model does not take advantage of the noise removal procedure. 
Considering the fact that \cite{endo2016deep}'s model needs high computational power for training a CNN and the use of geographical information raises a question of which method performed better. 
Moreover, the assumption of knowing the test set labels for segmentation and image generation are factors that \cite{endo2016deep} took advantage of it, indirectly. 
Setting up an experiment that divides the training and test based on user and city can give a good answer to the above question which can be done in future work.

\section{Comparison with Dabiri et al.[8]}
The label set for \cite{dabiri2018inferring}'s research is walking, train, bus, bike, taxi, subway and car so that the taxi and car are merged and called driving. 
Moreover, subway and train merged and called train. 
We filtered the Geolife data to get the same subsets as \cite{dabiri2018inferring} reported based on that. 
Then, we randomly selected 80\% of the data as the training and the rest as test set- we applied five fold cross validation.

The best subset of features were applied same as the previous experiment.
Running the random forest classifier with 50 estimators, using SKlearn implementation \cite{scikit-learn}, gives a mean accuracy of 88.5\% for the five fold cross validation. 
Comparing the results of the cross validation with the reported accuracy of \cite{dabiri2018inferring}'s research, we checked the normality of cross validation results and applied the appropriate statistical test. 
Therefore, we ran one-sample wilcoxon-test to compare the results with the reported result of \cite{dabiri2018inferring}, 84.8\%. 

A one sample Wilcoxon Signed-ranks test indicated that our accuracy results (88.50\%) is higher than Dabiri et al.\cite{dabiri2018inferring}'s results (84.8\%), p=0.0796. 


We avoided using the noise removal method in the above experiment because we believe we do not have access to labels of test dataset and using this method only increases our accuracy unrealistically. 
However, we explored the noise removal separately to show how much this procedure changes the accuracy of our model.

Dabiri et al. \cite{dabiri2018inferring} applied a noise removal method using the ground truth. 
They chose an upper bound and a lower bound for each transportation mode in their pre-processing step. 
They removed samples which were out of the predefined bounds. 
We found the boundaries reported in table \ref{tab:bounds}. 
Then we removed the out of bound samples and trained a random forest classifier. 
The results show that this method increases the mean accuracy from 88.5\% to 91.8\%. 
This improvement relies on the fact that we know the class labels of test set in the pre-processing step. 
The question we try to solve is predicting the transportation modes; therefore, this assumption is not valid.

\section{Cross validation methods}
In this research, three different methods of cross validation were observed, cross validation by dividing users\cite{endo2016deep}, cross validation by including users\cite{zheng2010understanding}, and random cross validation\cite{dabiri2018inferring,xiao2017identifying,liu2017end,zhu2018transportation}. 
The first and last methods were implemented to study whether there is a statistically significant difference between correlation among cross validation folds in these two methods or not. 
First, a subset of the Geolife Dataset including car, bus, bike, and walk were selected for this experiment.
Note that we do not train a model in this experiment and we only compared correlation between folds in these two methods of cross validation.

In \cite{STDM2017} and \cite{roberts2017cross}, the authors explained that the samples in spatio-temporal data are correlated. This relationship in time and space can cause optimistic evaluation of classifiers\cite{roberts2017cross}.
In this experiment, data was divided into five folds using random method (random) and dividing by user (user oriented) methods. 
Then, the spearman's rank-order correlation between all trajectory features was computed. 
Then, a table with two columns, random method and user oriented method of calculated Spearman's rank-order correlations were created. 
The Spearman's rank-order correlations mean for random method is 0.303 and the mean for user oriented method is 0.251. 
Both methods suffer from linear correlation between the training samples and test samples; however, is there any significant difference between these two methods?

We needed to select a non parametric method to compare the correlation data of the two methods; so, the Mann-Whitney rank test was chosen.

\begin{figure}[ht]
\caption{the difference between \textit{user oriented cross-validation} and \textit{random cross-validation}}
\centering
\includegraphics[width=0.8\textwidth]{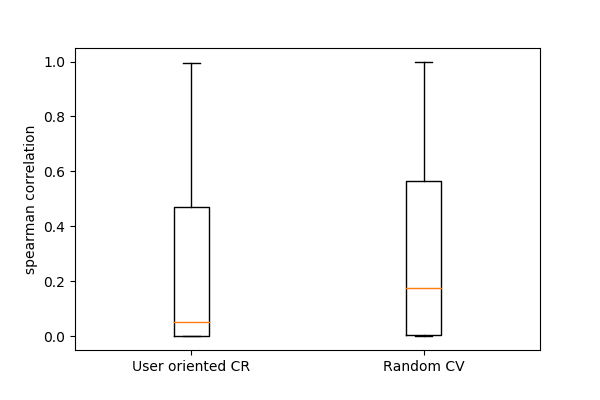}
\label{fig:spearman}
\end{figure}

The Mann-Whitney rank test between \textit{random method} correlations and \textit{user oriented method} correlations were calculated and the result shows that the null hypothesis is rejected which means both samples are NOT from the same distribution. 
Thus, there is a statistically significant difference between the median of these two methods(statistic=550680.0, p-value=1.95e-12). 
Figure \ref{fig:spearman} shows the difference between the \textit{random method} and the \textit{user oriented method}. 
The results of this experiment suggest that using user oriented cross-validation provides less correlation between the training set and test set. 
The non-linear correlation between the training and test set can be investigated as future studies. 

This experiment can be continued by doing an experiment on the heterogeneity of divided sets as a future work.

\begin{figure}[ht]
\caption{the difference cross validation results for \textit{user oriented cross-validation} and \textit{random cross-validation}}
\centering
\includegraphics[width=0.8\textwidth]{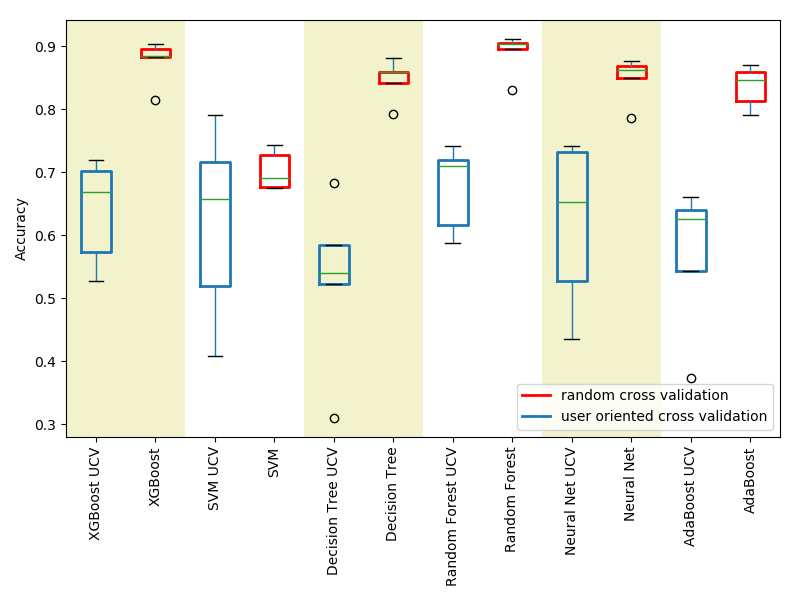}
\label{fig:clfs_ucv}
\end{figure}

In order to visualize the effect of type of cross validation on transportation modes prediction task, we setup a controlled experiment. We use same classifiers and same features to calculate the cross validation accuracy. Only the type of cross validation is different in this experiment, one is random cross validation and another is user oriented cross validation. Figure \ref{fig:clfs_ucv} shows the results of cross validation for different classifiers. The way random cross validation is optimistic is clearly shown here.

Figure \ref{fig:clfs_ucv} shows that there is a considerable difference between the cross validation results of user oriented cross validation and random cross validation. 
This graph indicates that random cross validation provides optimistic accuracy results. 
Since the correlation between user oriented cross validation results is less than random cross validation, proposing a specific cross validation method for evaluating the transportation mode prediction is a topic that needs attention.
\chapter{Conclusions}
\label{ch:conclusions}



\section{Conclusions and Future Works}

In this research, we reviewed some recent transportation modes prediction methods and feature selection methods. 
The framework proposed in \cite{etemad2018predicting} for transportation modes prediction was extended and  five experiments were conducted to cover different aspects of transportation modes prediction. 

First,the performance of six recently used classifiers for the transportation modes prediction was evaluated. 
The experiment was conducted using labels applied in \cite{dabiri2018inferring} and the same cross validation method. 
The results show that the random forest classifier performs the best among all the evaluated classifiers.
The SVM was the worst classifier and the accuracy result of XGBoost was competitive to the random forest classifier.

In the second experiment, the effect of features using two different approaches, wrapper method and information theoretical method were evaluated.
The wrapper method shows that we can achieve the highest accuracy using the top 20 features. Both approaches suggest that the  $F_{p90}^{speed}$ (the percentile 90 of the speed as defined in section \ref{sec:trajfeatures}) is the most important feature among all 70 introduced features. 
This feature is robust to noise since the outlier values do not contribute in the calculation of percentile 90.

In the third experiment, the best model was compared with the results showed in  \cite{endo2016deep}. The results show that our suggested model achieved a higher accuracy. Our applied features are readable and interpretable in comparison to \cite{endo2016deep} and our model has less computational cost.

In the forth experiment, the best model is compared with \cite{dabiri2018inferring}. The accuracy results show that our model achieved higher accuracy. 
The effect of the ground truth noise removal method was investigated. The cleaned dataset achieved a higher accuracy. However, this achievement is optimistic since we do not have access to the test set labels in the pre-processing step.

In the last experiment,the correlation between the training set and test set in two different cross validation approaches were compared (e.g., user oriented cross validation and random cross validation). 
The result shows that there is a statistically significant difference between the correlation among the training and test set of each approach.

\subsection{Future Work}
The segmentation method plays an important role in the transportation modes prediction task. 
Some segmentation methods use stop behaviour as a metric to segment trajectories. 
Some algorithms use a fixed time window and some methods apply other features related to trajectory to generate semantic trajectory segmentation.
It can be a promising future work to evaluate available segmentation methods and possibly propose a new method. 

Privacy preserving methods for using raw trajectories is another important aspect of this research. C-safety is an example of an anonymization framework for semantic trajectory\cite{monreale2011c}. Providing methods that can protect privacy of users are very important since some recent research identify users by using their raw trajectory or semantic trajectory\cite{furletti2012identifying}.

Filling the missing gaps in each trajectory can be another subject to research in future. Some data-sources such as S-AIS or the Geolife, when transportation happening underground, have missing gaps due to loss of their access to the satellites.
Moreover, trajectory classification can be used to detect anomalous behaviours in trajectories. For example, detecting an illegal immigration behaviour happening on the sea. 
Some interpolation methods such as Constraint random walk (for animal movement)\cite{technitis2015b}, cubic spline and cubic hermite interpolation (for AIS data)\cite{zhang2017enhance}, catmull-rom (applied in computer graphics)\cite{barry1988recursive}, Bezier curves( for animal tracking in a fluid environment)\cite{tremblay2006interpolation}, and kinematic interpolation( for transportation modes) \cite{long2016kinematic} have been introduced.

Furthermore, proposing a method for cross validation and feature selection for trajectory data is another important future work. 
When we have hierarchical structure, auto correlation or heterogeneity in data the conventional cross validation methods provide results optimistically.
David et al. investigated the issue of using cross validation in cases there is hierarchical structure, temporal or spatial dependencies\cite{roberts2017cross}. 
They showed that the cross validation in these cases are optimistic; however, they did not investigate the trajectory data structure.
proposing algorithms that consider these properties of data is very helpful to have more accurate evaluation of data.

\chapter{Appendices}

\appendix
\newpage
\section{Appendices}

\subsection{Statistical tests  appendix}

In this work, some statistical tests are used that we review them in the following and before applying them.
\subsubsection{Shapiro Wilk test}
The \textit{Shapiro Wilk test} applies for checking the abnormality of data\cite{shapiro1965analysis}. 
The \textbf{null hypothesis} is \textit{the observations came from a normally distributed population}. 
The goal is to reject the null hypothesis. 
The Python SKlearn implementation of the Shapiro test, \textit{scipy.stats.shapiro}, was used in this research\cite{scikit-learn}.
The documentation of the SKlearn emphases that for N greater than 5000 the statistic is accurate but the p-value may not be accurate. 
Moreover, the probability of false positive or type I error for this test is close to 5\%\cite{scikit-learn}. 
The analysis of p-values is as follows.

If the \textit{p-value} is \textit{less than} $\alpha$, we reject the null hypothesis which means the observations do not come from a normally distributed population.

If the \textit{p-value} is greater than $\alpha$, we fail to reject the null hypothesis. 
It does not mean that we can accept the null hypothesis.

\subsubsection{Kolmogorov-Smirnov test}
The Kolmogorov-Smirnov test for goodness of fit is a non parametric test to evaluate the normality of the observations.

The \textbf{null hypothesis} is \textit{the distribution of the two observations are identical.}\cite{scikit-learn}. 
To check the normality of a distribution, we use normal distribution as one of the observations. 
Therefore, we call the function using \textit{stats.kstest(observation,'norm', alternative = 'two-sided')}.
The alternative hypothesis can be set to \textit{less} or \textit{greater} for comparing two observations.

If the \textit{p-value} is \textit{less than} $\alpha$, we reject the null hypothesis which means the observations distributions are not identical.

If the \textit{p-value} is greater than $\alpha$, we fail to reject the null hypothesis.

\subsubsection{Wilcoxon rank-sum test}
The non-parametric wilcoxon rank-sum statistic for two samples is a non parametric test that compare two related observation. It usually uses as alternative non-parametric test for paired student t-test.
The \textbf{null hypothesis} is \textit{the two observations come from the same distribution.}\cite{scikit-learn}.
we use the sklearn implementation of this test using \textit{scipy.stats.ranksums}\cite{scikit-learn}.

If the \textit{p-value} is \textit{less than} $\alpha$, we reject the null hypothesis which means the observations come from two different distributions.

If the \textit{p-value} is greater than $\alpha$, we fail to reject the null hypothesis. 

\subsubsection{Mann-Whitney test}

The \textit{Mann-Whitney test} is a non-parametric to compare two independent observations. This is an alternative test for independent t-test when the observations distributions are asymmetric or is not following the normal distribution.
The \textbf{null hypothesis} that \textit{the distributions of the two samples are equal}. 
The alternative hypotheses is that the medians of the two groups are not equal.

If the P value is less than $\alpha$, the null hypothesis can be rejected. It means the medians of the two samples are not equal.

If the P value is greater than $\alpha$, cannot reject the null hypothesis. therefore, the data do not give you any reason to reject the null hypothesis. 
\subsubsection{One sample wilcoxon test}
The \textit{one sample wilcoxon test} is a non-parametric equivalent for one sample t-test. 
It compares an observation with one standard mean.
The null hypothesis and interpretation is same as Mann-Whitney test.
We use Sklearn to calculate One sample wilcoxon test by using \textit{scipy.stats.wilcoxon(observations-standardmean)}\cite{scikit-learn}.

\subsubsection{kruskal wallis test}
The \textit{kruskal wallis test} is a non-parametric test to compare two or more than two samples. 
The \textit{null hypothesis} is that \textit{the distribution of two samples are equal}. 
The alternative hypotheses is that one of the distributions is shifted and they are not equal anymore.

If the P value is less than $\alpha$, the null hypothesis can be rejected. It means one of the samples is different from the other samples.

If the P value is greater than $\alpha$, we fail to reject the null hypothesis. 
Therefore, samples come from same distribution.

\newpage

\subsection{Parameters of the applied classifiers} 
\label{sec:param}

\begin{table}[ht]
\caption{parameters of the applied classifiers}
\label{tab:param}
\begin{tabular}{|l|l|}
\hline
\multicolumn{1}{|c|}{\textbf{Classifier}} & \multicolumn{1}{c|}{\textbf{parameters}}                                                                                                                                    \\ \hline
XGBoost  &\begin{tabular}[c]{@{}l@{}}criterion=friedman\_mse, init=None, learning rate=0.1, loss=deviance,\\ max depth=3, max features=None, max leaf nodes=None, min \\ impurity decrease=0.0, min impurity split=None, min samples leaf=1, min \\ samples split=2, min weight fraction leaf=0.0, n\_estimators=100,\\ presort=auto, subsample=1.0, verbose=0, warm start=False \end{tabular} \\ \hline
SVM  &  \begin{tabular}[c]{@{}l@{}}C=1.0, cache size=200, class weight=None, coef0=0.0, decision function\\ shape=ovr, degree=3, gamma=auto, kernel=rbf, max iter=-1,\\ probability=False, shrinking=True, tol=0.001, verbose=False\end{tabular}\\ \hline
decision tree &
\begin{tabular}[c]{@{}l@{}}class weight=None, criterion=gini, max depth=5, max features=None,\\ max leaf nodes=None, min impurity decrease=0.0, min impurity \\split=None, min samples leaf=1, min samples split=2, min weight\\ fraction leaf=0.0, presort=False, splitter=best\end{tabular}\\ \hline
random forest&\begin{tabular}[c]{@{}l@{}}bootstrap=True, class weight=None, criterion=gini, max depth=None,\\ max features=auto, max leaf nodes=None, min impurity decrease=0.0,\\ min impurity split=None, min samples leaf=1, min samples split=2, \\min weight fraction leaf=0.0, n\_estimators=50, n\_jobs=1, oob\_score=False,\\ verbose=0, warm start=False\end{tabular}\\ \hline
neural network&  \begin{tabular}[c]{@{}l@{}}activation=relu, alpha=1e-09, batch size=auto, beta\_1=0.9, beta\_2=0.999,\\ early stopping=False, epsilon=1e-08, hidden layer sizes=(140,),\\ learning\_rate=constant, learning\_rate\_init=0.01, max iter=1000,\\ momentum=0.9, nesterovs momentum=True, power\_t=0.5, shuffle=True, \\solver=adam, tol=1e-09, validation fraction=0.1, verbose=0,\\ warm start=False\end{tabular} \\ \hline
adaboost& \begin{tabular}[c]{@{}l@{}}algorithm=SAMME.R, base estimator=None,\\ learning\_rate=1.0, n\_estimators=50\end{tabular} \\\hline
\end{tabular}
\end{table}

\newpage
\bibliographystyle{plain}
\bibliography{ref}

\end{document}